\documentclass[conference]{IEEEtran}
\IEEEoverridecommandlockouts

\usepackage{graphicx}
\usepackage{wrapfig}
\usepackage{microtype}
\usepackage{hyperref}
\usepackage{url}
\usepackage{booktabs}

\usepackage{amsmath}
\usepackage{amsfonts}
\usepackage{graphicx}
\usepackage{amssymb}
\usepackage{booktabs} 
\usepackage{musicography}
\usepackage{multirow}
\usepackage[inline]{enumitem} 
\usepackage{graphicx}
\usepackage{subfigure}
\usepackage{xcolor}
\definecolor{LightSteelBlue1}{RGB}{202,225,255}
\definecolor{LightPink}{RGB}{245,191,210}
\definecolor{Moccasin}{RGB}{255, 228, 181}
\usepackage{tikz}
\usepackage{pifont}
\usepackage{xspace}
\usepackage{bm}
\usepackage{tablefootnote}

\usepackage{multirow}

\newcommand\blfootnote[1]{%
  \begingroup
  \renewcommand\thefootnote{}\footnote{#1}%
  \addtocounter{footnote}{-1}%
  \endgroup
}

\definecolor{LightSteelBlue1}{RGB}{202,225,255}

\def\figref#1{Figure~\ref{fig:#1}}
\def\figlabel#1{\label{fig:#1}\label{p:#1}}

\usepackage{cite}
\usepackage{amsmath,amssymb,amsfonts}
\usepackage{algorithmic}
\usepackage{graphicx}
\usepackage{textcomp}
\usepackage{xcolor}
\usepackage{wrapfig}
\def\BibTeX{{\rm B\kern-.05em{\sc i\kern-.025em b}\kern-.08em
    T\kern-.1667em\lower.7ex\hbox{E}\kern-.125emX}}
\begin{document}

\title{Why Lift so Heavy? Slimming Large Language Models by Cutting Off the Layers
}

\author{Shuzhou Yuan$^\ast$$^{1}$,~Ercong Nie$^\ast$$^{2,3}$,~Bolei Ma$^{2,3}$,~Michael Färber$^{1}$ \\
$^{1}$ScaDS.AI and TU Dresden 
$^{2}$LMU Munich \\
$^{3}$Munich Center for Machine Learning (MCML) \\
\texttt{shuzhou.yuan@tu-dresden.de,~nie@cis.lmu.de,~bolei.ma@lmu.de}}

\maketitle

\begin{abstract} Large Language Models (LLMs) demonstrate exceptional language understanding and generation capabilities by learning from context. Leveraging the strong in-context learning (ICL) abilities of LLMs, prompt-based fine-tuning has proven to be effective for enhancing the adaptability and alignment of LLMs, especially in low-data scenarios. However, the billions of parameters resulting from layer stacking in LLMs present significant computational challenges, limiting the practicality of fine-tuning. 
To tackle this problem, we explore the application of layer-wise model pruning in prompt-based fine-tuning of LLMs for few-shot learning scenarios. Our approach involves dropping certain model layers and fine-tuning the model with the remaining layers.
Surprisingly, we observe that even with fewer layers, LLMs maintain similar or better performance levels, particularly in prompt-based fine-tuning for text classification tasks. Remarkably, in certain cases, models with a single layer outperform their fully layered counterparts. These findings offer valuable insights for future work aimed at mitigating the size constraints of LLMs while preserving their performance, thereby opening avenues for significantly more efficient use of LLMs.\blfootnote{$^\ast$ Equal contribution.}

\end{abstract}


\section{Introduction}

Large Language Models (LLMs) have revolutionized natural language processing (NLP), achieving state-of-the-art performance across diverse tasks such as text classification and language generation. 
Their success stems from deep architectures with stacked transformer layers~\cite{vaswani2017attention}, enabling rich text representations and fluent outputs.
Since the original Transformer with six decoder layers~\cite{vaswani2017attention}, LLMs have rapidly evolved, exemplified by models like Llama2 7B with 32 layers~\cite{touvron2023llama}, OPT with 24 layers~\cite{zhang2022opt}, and GPT2-XL with 48 layers~\cite{radford2019language}.
However, this architectural complexity comes at a significant cost: the computational demands for fine-tuning, storage, and inference grow exponentially with model size, making LLMs increasingly impractical for resource-constrained applications.

To address these challenges, researchers have explored various strategies to reduce the training and inference costs of LLMs. \emph{Parameter-efficient fine-tuning (PEFT)} methods, such as Adapter Tuning~\cite{pmlr-v97-houlsby19a}, Prompt Tuning~\cite{lester-etal-2021-power}, and Prefix Tuning~\cite{li-liang-2021-prefix}, freeze most of the model's parameters and introduce lightweight trainable modules. 
While these methods reduce training costs, they do not address the fundamental issue of the model's size, leaving storage and inference inefficiencies unresolved. Another line of research focuses on \emph{model pruning}, which removes non-essential weights or layers to compress the model~\cite{Han2015DeepCC,sanh2020movement,xu-etal-2021-rethinking}.

\begin{figure}[t]
    \centering
    \includegraphics[width=0.7\linewidth]{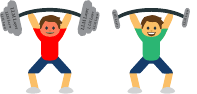}
    \caption{Our research demonstrates that retaining only a single layer in LLMs, by simply cutting off additional layers, sustains equivalent performance compared to utilizing the full complement of layers.}
    \label{intro_figure}
\end{figure}

In this work, we propose a simple yet effective approach to address the size and efficiency challenges of LLMs: \emph{layer-wise pruning combined with prompt-based fine-tuning}. Specifically, we investigate the impact of removing top layers from decoder-only LLMs and fine-tuning the remaining layers using prompt-based methods. 
Prompt-based fine-tuning has emerged as a powerful paradigm for adapting LLMs to downstream tasks, particularly in low-data scenarios, by leveraging the model's in-context learning (ICL) capabilities~\cite{brown2020language,gao-etal-2021-making,wei2022emergent}. By combining layer pruning with prompt-based fine-tuning, we aim to explore whether LLMs can maintain their performance with significantly fewer layers, thereby reducing both storage and computational costs.

Through extensive experiments on text classification tasks, we observed that LLMs with drastically reduced layers can achieve comparable or even superior performance to their full-layer counterparts. For instance, we demonstrate that a single-layer version of GPT-2 XL~\cite{radford2019language} achieves performance on par with the full 48-layer model, while reducing the parameter count by over 93\%. Similarly, we observe consistent trends with the OPT model~\cite{zhang2022opt}, where significant layer reductions result in minimal performance degradation. 
These findings suggest that deeper models may not always be necessary for optimal performance and open new avenues for efficient LLM deployment.

\begin{figure*}[t]
	\centering  
	\subfigure[\text{Vanilla fine-tuning}]{  
		\begin{minipage}{.482\linewidth}
			\centering    
			\includegraphics[width=\linewidth]{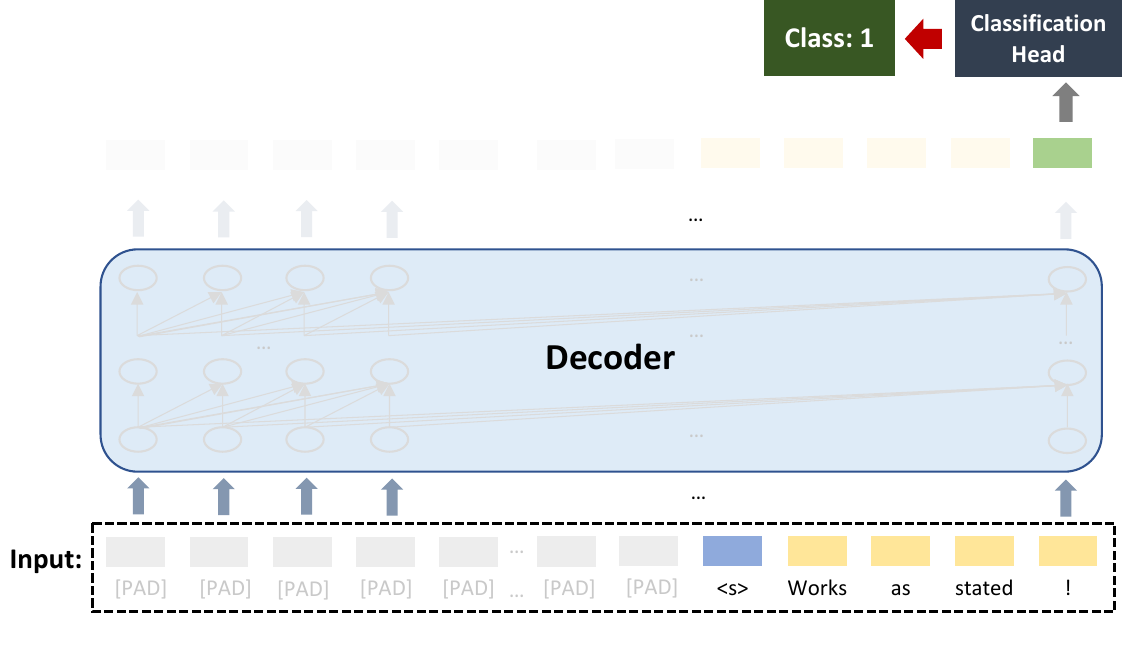} 
		\end{minipage}
	}
        \subfigure[\text{Prompt-based fine-tuning}]{ 
		\begin{minipage}{.482\linewidth}
			\centering    
			\includegraphics[width=\linewidth]{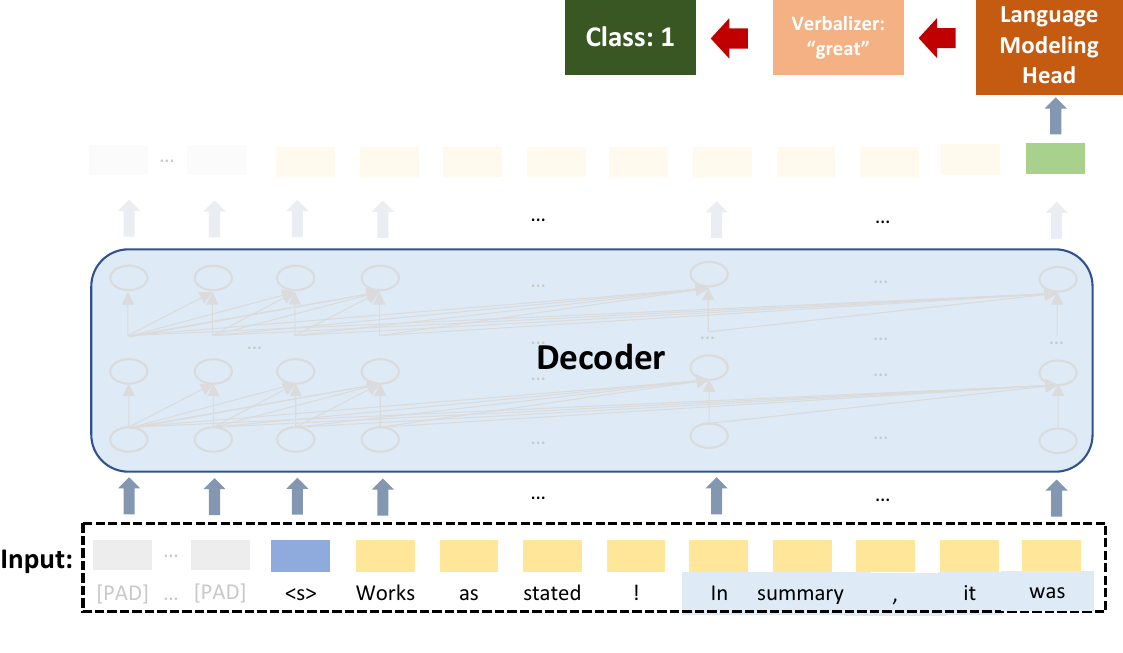}
		\end{minipage}
	}
\caption{The comparison of vanilla fine-tuning and prompt-based fine-tuning. $<s>$ (blue) and [PAD] (grey) are special tokens. Tokens with a blue background in the input text refer to the prompt template. Green block indicates the hiddle state of the last input token. In the vanilla fine-tuning (left), it can be used to predict the label via the classification head; in the prompt-based fine-tuning (right), it can be used to generate the label word via outputting the next token through the language modeling head. The verbalizer is a function mapping from the task label set to a subset of the model vocabulary.}    
    \figlabel{fig_comparison}   
\end{figure*}

Our contributions are summarized as follows:

\begin{enumerate}[label={\textbf{\roman{*})}}]
\item We introduce a simple layer-wise pruning strategy for LLMs and evaluate its effectiveness in prompt-based fine-tuning for few-shot learning scenarios.

\item We systematically analyze the impact of layer pruning on LLM performance across multiple classification tasks, revealing that significant parameter reductions are possible without compromising accuracy.
	
\item Our findings provide practical insights into the trade-offs between model size and performance, offering a promising direction for reducing the computational and storage costs of LLMs.

\end{enumerate}

\begin{table*}[t]
\centering
\caption{Template for prompt. $[S]$ and $[L]$ refer to the input and label word of the demonstration. $[S_i]$ refers to the input test sentence.}\label{prompt_temp}
\scalebox{0.85}{
\begin{tabular}{l|l|l} 
\toprule
\textbf{Task} & \textbf{Template~}                                                                                                                                                              & \textbf{Label Words}                                                                                   \\ 
\midrule
AGNews        & \begin{tabular}[c]{@{}l@{}}\texttt{Article: $[S]$ Answer: $[L]$ Article: $[S_i]$ Answer:~\_\_\_}\end{tabular}             & \begin{tabular}[c]{@{}l@{}}World, Sports,\\ Business, Technology\end{tabular}                          \\
\midrule
EmoC          & \begin{tabular}[c]{@{}l@{}}\texttt{Dialogue: $[S]$ Emotion: $[L]$ Dialogue:$[S_i]$ Emotion:~\_\_\_}\end{tabular}           & \begin{tabular}[c]{@{}l@{}}Happy, Sad,\\Angry, Others\end{tabular}                                    \\ 
\midrule
SST-2         & \begin{tabular}[c]{@{}l@{}}\texttt{Review: $[S]$ Sentiment: $[L]$ Review: $[S_i]$ Sentiment:~\_\_\_}\end{tabular}              & Positive, Negative                                                                                     \\ 
\midrule
TREC          & \begin{tabular}[c]{@{}l@{}}\texttt{Question: $[S]$ Answer Type: $[L]$ Question: $[S_i]$ Answer Type:~\_\_\_}\end{tabular} & \begin{tabular}[c]{@{}l@{}}Abbreviation, Entity, \\Description, Person, \\Location, Number\end{tabular}  \\
\bottomrule
\end{tabular}}
\label{temp}
\end{table*}

\section{Related Work}
\subsection{Model Pruning}
Pruning, a prevalent technique for model compression method, seeks to discard non-essential parameters within language models, thereby reducing computational and storage costs while maintaining performance~\cite{lecun1989optimal,xia-etal-2022-structured,wang-etal-2023-distill}. 
Pruning methods can be broadly categorized into \emph{unstructured pruning}, which removes individual weights, and \emph{structured pruning}, which removes higher-granularity structures such as neurons, attention heads, or entire layers.
\emph{Layer-wise pruning}, a specific form of structured pruning, has been explored in several studies.
Fan et al. (2019) \cite{fan2019reducing} introduced LayerDrop, a structured dropout method that randomly drops layers during training to improve robustness to layer reduction at inference time.
Additionally, several studies have investigated the disparities in representations across different layers, highlighting the significance of updating the last few layers for language models \cite{kovaleva-etal-2019-revealing,liu-etal-2019-linguistic,lee2019would}.
Building on this foundation, Peer et al. (2022) \cite{peer2022greedy} and Sajjad et al. (2023) \cite{SAJJAD2023101429} aimed to identify and remove an optimal subset of layers directly from the pretrained models for use in downstream tasks.
These works focus on pruning layers to reduce model size while preserving task-specific performance within the standard fine-tuning paradigm. However, they do not explore the interaction between layer pruning and prompt-based fine-tuning, which is the focus of our work.

Recent studies have explored the application of pruning techniques to large language models, such as Sheared LLMs~\cite{xia2023sheared} and SparseGPT~\cite{frantar2023sparsegpt}, which aims to sparsify or prune LLMs for efficient inference. 
These methods often involve complex optimization strategies or require additional training steps~\cite{jha2023train,chen2023lorashear,ashkboos2024slicegpt}. In contrast, our approach is a simple top-layer pruning strategy that directly removes layers from LLMs and evaluates their performance in prompt-based fine-tuning scenarios. Unlike prior work, we quantitatively investigate the impact of layer pruning on LLMs in few-shot learning settings, demonstrating that even drastic reductions in layers can maintain or improve performance.

\subsection{Prompt-based Fine-tuning}
Prompt-based learning has emerged as a powerful paradigm for adapting pretrained language models to downstream tasks.
Brown et al.\cite{brown2020language} demonstrated the effectiveness of in-context learning (ICL), where task-specific prompts are used to guide the model without updating its parameters. 
Large-scale pretrained language models such as GPT-3 can learn to perform a language understanding and generation task from task descriptions and demonstrations by putting examples of input-output pairs into the input as a context. 
This in-context learning (ICL) paradigm is a type of prompt-based learning method. 
Building on this, prompt-based fine-tuning has been proposed as a method to align the fine-tuning objective with the pretraining task~\cite{gao-etal-2021-making,schick-schutze-2021-exploiting}. 
Prompt-based fine-tuning uses natural language patterns to reformulate input examples into cloze-style phrases, as illustrated in \figref{fig_comparison}b. This conversion aligns the fine-tuning objective with the language modeling task. Specifically, encoder models involve mask token prediction, while for decoder models, the objective is next token prediction.
This approach has shown particular promise in low-data scenarios, where it can outperform traditional fine-tuning methods~\cite{schick-schutze-2021-just}.
Recent advancements in prompt-based fine-tuning have extended its applicability to multilingual tasks~\cite{ma-etal-2023-prompt}, sequence labeling~\cite{ma-etal-2024-topro}, and parameter-efficient fine-tuning methods\cite{yuan2024lgnnavi}. 

Our work bridges the gap between model pruning and prompt-based fine-tuning by investigating how layer-wise pruning affects the performance of LLMs in prompt-based fine-tuning scenarios. While prior studies have explored parameter-efficient fine-tuning and layer pruning independently, we are the first to systematically evaluate the interaction between these two techniques, providing insights into how LLMs can be slimmed down for efficient few-shot learning.

\section{Problem Statement}
In this work, we explore a simple yet effective approach to address the size and efficiency challenges of LLMs: \emph{layer-wise pruning combined with prompt-based fine-tuning}. Specifically, we investigate whether all layers in decoder-only LLMs are necessary for downstream tasks, particularly in \emph{few-shot learning} scenarios. 
Our approach involves removing the top $k$ decoder layers of a pretrained LLM and fine-tuning the remaining layers using prompt-based methods. This top-layer dropping strategy is motivated by the observation that lower layers in transformer architectures often capture general linguistic features, while upper layers specialize in task-specific representations~\cite{kovaleva-etal-2019-revealing,liu-etal-2019-linguistic}. By pruning the top layers, we aim to reduce the model size while retaining its core capabilities.

\subsection{Model Layer Dropping}
Formally, let $\mathcal{M}$ be a decoder-only pretrained language model consisting of an embedding layer $\mathcal{E}_0$, $n$ decoder layers with identical architecture: $\{l_1,...,l_n\}$, and a language modeling head $l_m$. 
In the \emph{top-layer dropping strategy}, we remove the top $k$ decoder layers $\{l_{n-k+1},\cdots,l_n\}$, leaving the reduced model with $n-k$ layers. The resulting model consists of the embedding layer $\mathcal{E}_0$, the remaining decoder layers $\{l_1,\cdots,l_{n-k}\}$, and the language modeling head $l_m$.
The reduced model is then fine-tuned on downstream tasks using prompt-based methods.

\begin{table*}[t]
\centering
\caption{Prompt-based fine-tuning with different numbers of layers using language modeling head. $n$ denotes the number of layers.}
\scalebox{.85}{
\begin{tabular}{c|ccccccc} 
\toprule
 & \textbf{$n$} & \textbf{Param} & \textbf{AGNews} & \textbf{EmoC}  & \textbf{SST-2} & \textbf{TREC}  & \textbf{Average} \\ 
\midrule                                                                         
\multirow{5}{*}{\textbf{GPT2-XL}} & 48              & 1.6B           & 85.34           & 73.70          & 68.97          & 80.16          & 77.04             \\
& 24              & 819M           & \textbf{86.66}  & 75.44          & 72.41          & 72.41          & 76.73            \\
& 12              & 450M           & 86.42           & 75.34          & 73.76          & 84.72          & 80.06           \\
& 2               & 143M          & 85.80           & 75.78          & \textbf{74.22} & \textbf{85.12} & \textbf{80.23}   \\
& 1               & 112M           & 85.30           & \textbf{76.76} & 72.89          & 83.16          & 79.53           \\
\midrule
\multirow{5}{*}{\textbf{OPT}} & 24              & 1.3B           & 76.10           & 74.80          & 64.31          & 76.80          & 73.00 \\
& 12              & 711M           & 77.52           & \textbf{78.46} & 69.13          & 80.44          & 76.39             \\
 & 6               & 409M           & 77.48           & 77.22          & \textbf{69.34} & 81.72          & 76.44             \\
 & 2               & 297M           & 80.60           & 74.66          & 70.92          & 81.88          & 77.01             \\
  & 1               & 157M           & \textbf{81.32}  & 77.42          & 69.17          & \textbf{82.12} & \textbf{77.51}    \\
\bottomrule
\end{tabular}}
\label{lm_results}
\end{table*}

\subsection{Task Formulation}

Our experiments focus on text classification tasks in few-shot learning scenarios, where the goal is to predict the correct class given a small number of labeled examples. We reformulate the classification task as a language modeling problem using prompt-based fine-tuning. 

Let $\mathcal{M}$ be a language model with vocabulary $V$, and let $\mathcal{L}$ be a set of label words. The training set $\mathcal{T}$ consists of pairs $(s, l)$, where $s$ is a sequence of tokens from the vocabulary $V$ and $l$ is a label word from the set $\mathcal{L}$.
To align the task with the language modeling objective, we define a pattern $\mathcal{P}(s, l)$ that maps an input text $s$ and its label $l$ into a cloze-style prompt.
For example, in an emotion detection task, the input text $s=\text{`I enjoyed it a lot!'}$ and a label word $l=\text{`Happy'}$ are reformulated as:
$$
\colorbox{LightSteelBlue1}{\text{Dialogue: \underline{I enjoyed it a lot!} Emotion: \underline{Happy}}}
$$

For a $k$-class classification task, we define a prompt $X(s)$ by concatenating several patterns $\mathcal{P}(s_i, l_i)$ with the input text $s$ to be classified:

\begin{equation}
\label{equa_prompt}
X(s) = \mathcal{P}(s_1, l_1)\oplus \ldots \oplus \mathcal{P}(s_k, l_k)\oplus \mathcal{P}(s, \varepsilon)
\end{equation}

Here, $\oplus$ denotes the concatenation of the input demonstrations, and $\varepsilon$ is the empty string. The model $\mathcal{M}$ predicts the next token \( l_i \), which is assigned as the label word for $s$. The model is initialized with pretrained parameters \( \phi \), and fine-tuned by minimizing the cross-entropy loss:
\begin{equation}
  \ell = -\sum_{(s, l) \in \mathcal{T}} \log p_\phi(X(s), l)
\end{equation}
where $p_\phi(X(s), l)$ is the probability assigned by $\mathcal{M}$ to the correct label $l$. To ensure a fair evaluation, we randomly select one demonstration per class to form the prompt and exclude these examples from the training set $\mathcal{T}$. The remaining examples in $\mathcal{T}$ are sampled for fine-tuning.

\section{Experiment}
\subsection{Dataset and Model}

We evaluate our proposed layer-dropping strategy on four widely used text classification datasets, which span diverse domains and task types:

\begin{itemize}
    \item \textbf{AGNews:} a news topic classification dataset with four labels~\cite{NIPS2015_250cf8b5}.
    \item \textbf{EmoC:} EmoContext, a four-label emotion classification dataset~\cite{chatterjee-etal-2019-semeval}.
    \item \textbf{SST-2:} Stanford Sentiment Treebank Binary, a sentiment analysis dataset with 2 labels \cite{socher-etal-2013-recursive}.
    \item \textbf{TREC:} Text REtrieval Conference Question Classification, question classification dataset with 6 labels \cite{li-roth-2002-learning,hovy-etal-2001-toward}.
\end{itemize}
For each dataset, we design task-specific prompt templates to align the classification tasks with the language modeling objective. 
These datasets were selected to provide a controlled environment for evaluating the impact of layer pruning on classification tasks. 
While these datasets are relatively simple compared to more complex generative or reasoning tasks, they allow us to systematically study the effects of layer reduction on model performance.
The prompt templates and the dataset labels are detailed in Table \ref{prompt_temp}.

We use two pre-trained decoder-only language models from Huggingface~\cite{wolf-etal-2020-transformers}: \textbf{GPT-2 XL}\footnote{\url{https://huggingface.co/openai-community/gpt2-xl}}, a 48-layer model with 1.6 billion parameters, and \textbf{OPT-1.3B}\footnote{\url{https://huggingface.co/facebook/opt-1.3b}}, a 24-layer model with 1.3 billion parameters. These models were chosen for their widespread use and availability, enabling us to evaluate the impact of layer pruning on LLMs.

\begin{table*}[t]
\centering
\caption{Vanilla fine-tuning with different numbers of layers using classification head. $n$ denotes the number of layers.}
\scalebox{.85}{
\begin{tabular}{c|ccccccc} 
\toprule
& \textbf{$n$} & \textbf{Param} & \textbf{AGNews} & \textbf{EmoC}  & \textbf{SST-2} & \textbf{TREC}  & \textbf{Average}  \\ 
\midrule
\multirow{5}{*}{\textbf{GPT2-XL}} & 48           & 1.6B           & 86.62           & 72.60          & 72.96          & 81.64          & 78.46            \\
 & 24           & 819M           & 86.02           & 75.42          & 73.05          & 84.40          & 79.72            \\
& 12           & 450M           & \textbf{86.96}  & \textbf{76.58} & \textbf{73.76} & \textbf{85.04} & \textbf{80.59}   \\
& 2            & 143M          & 86.64           & 74.32          & 72.94          & 84.28          & 79.54            \\
& 1            & 112M           & 85.56           & 73.66          & 73.30          & 81.12          & 78.41            \\
\midrule
\multirow{5}{*}{\textbf{OPT}} & 24           & 1.3B           & 78.94           & 76.02          & 68.35          & 81.92          & 76.31             \\
& 12           & 711M           & 83.30           & 75.12          & 67.02          & 81.72          & 76.79             \\
& 6            & 409M           & 83.82           & \textbf{78.22} & 70.85          & 82.88          & \textbf{78.94}    \\
& 2            & 297M           & \textbf{84.38}  & 76.54          & \textbf{70.94} & 82.44          & 78.58             \\
& 1            & 157M           & 81.86           & 77.80          & 70.87          & \textbf{83.76} & 78.57             \\
\bottomrule
\end{tabular}}
\label{cls_results}
\end{table*}

\begin{table*}[t]
\centering
\caption{Prompt-based fine-tuning with different numbers of layers using classification head. $n$ denotes the number of layers.}
\scalebox{.85}{
\begin{tabular}{c|ccccccc} 
\toprule
& \textbf{$n$} & \textbf{Param} & \textbf{AGNews} & \textbf{EmoC}  & \textbf{SST-2} & \textbf{TREC}  & \textbf{Average} \\ 
\midrule

\multirow{5}{*}{\textbf{GPT2-XL}} & 48           & 1.6B           & 85.48           & 73.34          & 67.13          & 81.04          & 76.75            \\
& 24           & 819M           & 85.94           & 71.94          & 71.40          & 82.68          & 77.99           \\
& 12           & 450M           & \textbf{86.06}  & 73.82          & 72.18          & 83.28          & 78.83            \\
& 2            & 143M          & 85.28           & \textbf{75.52} & \textbf{73.19} & \textbf{84.12} & \textbf{79.53}   \\
& 1            & 112M           & 85.50           & 75.28          & 73.12          & 83.40          & 79.33            \\
\midrule
\multirow{5}{*}{\textbf{OPT}} & 24           & 1.3B           & 73.96           & 76.72          & 64.08          & 75.92          & 72.67             \\
 & 12           & 711M           & 76.02           & 75.86          & 68.60          & 80.72          & 75.30             \\
 & 6            & 409M           & 78.56           & \textbf{77.56} & \textbf{71.15} & 80.52          & \textbf{76.95}    \\
 & 2            & 297M           & 77.92           & 72.76          & 69.38          & 80.80          & 75.22             \\
 & 1            & 157M           & \textbf{79.88}  & 72.38          & 71.10          & \textbf{82.56} & 76.48             \\
\bottomrule
\end{tabular}}
\label{cls_prompt_results}
\end{table*}

\subsection{Experimental Setup\label{setting}}

To evaluate the impact of layer pruning, we experiment with models retaining different numbers of layers.
For GPT-2 XL, we retain 1, 2, 12, and 24 layers out of the original 48 layers, while for OPT-1.3B, we retain 1, 2, 6, and 12 layers out of the original 24 layers\footnote{The complete GPT-2 XL model consists of 48 layers, and the OPT-1.3b model comprises 24 layers.}.
For training, we set up a few-shot learning scenario by randomly sampling 200 training examples from the original training set of each dataset. 
An additional 1,000 samples from the original train set are used for validation, and 1,000 samples from the original test set (or the full test set, if smaller) are used for evaluation. 
To ensure fair evaluation, we randomly select one demonstration per class to construct the prompt by appending the sample to be predicted at the end of the demonstrations. 
These demonstrations are drawn from the original training set of the datasets and are excluded from subsequent training. 
Model selection is based on the accuracy achieved on the validation set, with the best-performing model then evaluated on the test set.

We conduct experiments using two fine-tuning paradigms: prompt-based fine-tuning and vanilla fine-tuning. In prompt-based fine-tuning, the task is reformulated as a language modeling problem, where the model predicts the label word as the next token. In vanilla fine-tuning, the task is treated as a standard classification problem, where the model predicts the label using a classification head. Additionally, we explore the effect of replacing the language modeling head with a classification head in prompt-based fine-tuning to isolate the impact of the head type on model performance across different layer configurations.

We maintain consistent hyperparameters across models, regardless of the number of layers. Specifically, all models are fine-tuned using the AdamW optimizer~\cite{loshchilov2018decoupled} with a learning rate of 5e-5. Training is conducted for 50 epochs with early stopping after 15 epochs based on validation accuracy. Due to resource constraints, we use a batch size of 1. Five random seeds ([0, 42, 421, 520, 1218]) are used, and the average accuracy across these seeds is reported. For evaluation, the average accuracy across these seeds is reported as the final result.

\section{Results and Analysis}
In this section, we analyze the experimental results of layer pruning in LLMs across different configurations and fine-tuning paradigms. We provide a detailed discussion of the observed trends, their implications, and potential explanations for the findings.

\subsection{Results of Prompt-based Fine-tuning}
The results of prompt-based fine-tuning are presented in Table \ref{lm_results}. 
For the GPT2-XL model, we observe that the model's performance remains remarkably stable across different layer configurations. When fine-tuned with the full 48 layers, the average accuracy across all datasets is 77.04\%. Reducing the model to 24 layers results in a negligible drop in accuracy to 76.73\%. Notably, the 2-layer configuration achieves the highest average accuracy of 80.23\%, outperforming the full-layer model. Even with a single layer, the model achieves an average accuracy of 79.53\%, indicating that significant parameter reductions (from 1.6 billion to 112 million parameters) can be achieved without compromising performance.

A similar trend is observed for OPT-1.3B. The full 24-layer model achieves an average accuracy of 73.00\%, while the single-layer configuration achieves the highest accuracy of 77.51\%. This corresponds to a parameter reduction from 1.3 billion to 157 million. Notably, the performance improves as the number of layers decreases, with the 2-layer configuration achieving 77.01\%. These results suggest that the upper layers of LLMs may introduce redundancy in classification tasks, while the lower layers capture sufficient linguistic and semantic information for effective task performance.

For topic classification tasks, such as AGNews and TREC, shallow models often achieve the highest accuracy. For AGNews, the single-layer OPT model achieves the highest accuracy of 81.32\%, while for TREC, the 2-layer GPT-2 XL model achieves the highest accuracy of 85.12\%, closely followed by the single-layer OPT model at 82.12\%. 
More complex sentiment analysis tasks exhibit slightly different trends compared to topic classification. For SST-2, deeper configurations perform slightly better. The 2-layer GPT-2 XL model achieves the highest accuracy of 74.22\%, while the 6-layer OPT model achieves the highest accuracy of 69.34\%. This suggests that sentiment analysis tasks may require slightly deeper representations compared to topic classification, likely due to the nuanced nature of sentiment and the need for more complex contextual understanding.
These results suggest that simpler tasks such as topic classification can be effectively handled by shallow models, as they primarily rely on general linguistic features. The redundancy in upper layers appears unnecessary for these tasks, making them well-suited for aggressive layer pruning.

\subsection{Results of Vanilla Fine-tuning}

We delve deeper into the classification task employing the conventional classification paradigm, wherein text categorization relies solely on the input text without demonstrations. By removing demonstrations and providing the model with input examples, we apply a classification head to categorize the text, adhering to the experimental settings articulated in \S\ref{setting}.
The results of vanilla fine-tuning using the classification head are presented in Table \ref{cls_results}. Similar to prompt-based fine-tuning, we observe that reducing the number of layers has minimal impact on performance. For GPT-2 XL, the full-layer model achieves an average accuracy of 78.46\%, while the 12-layer configuration achieves the highest accuracy of 80.59\%. The single-layer configuration achieves an accuracy of 78.41\%, comparable to the full-layer model. For OPT-1.3B, the full-layer model achieves an average accuracy of 76.31\%, while the 6-layer configuration achieves the highest accuracy of 78.94\%. The single-layer configuration achieves an accuracy of 78.57\%, again demonstrating that significant parameter reductions can be achieved without sacrificing performance.

\begin{figure}[h]
 \centering
 	\subfigure[\text{GPT2-XL}]{  
		\begin{minipage}{\linewidth}
			\centering    
			\includegraphics[width=\linewidth]{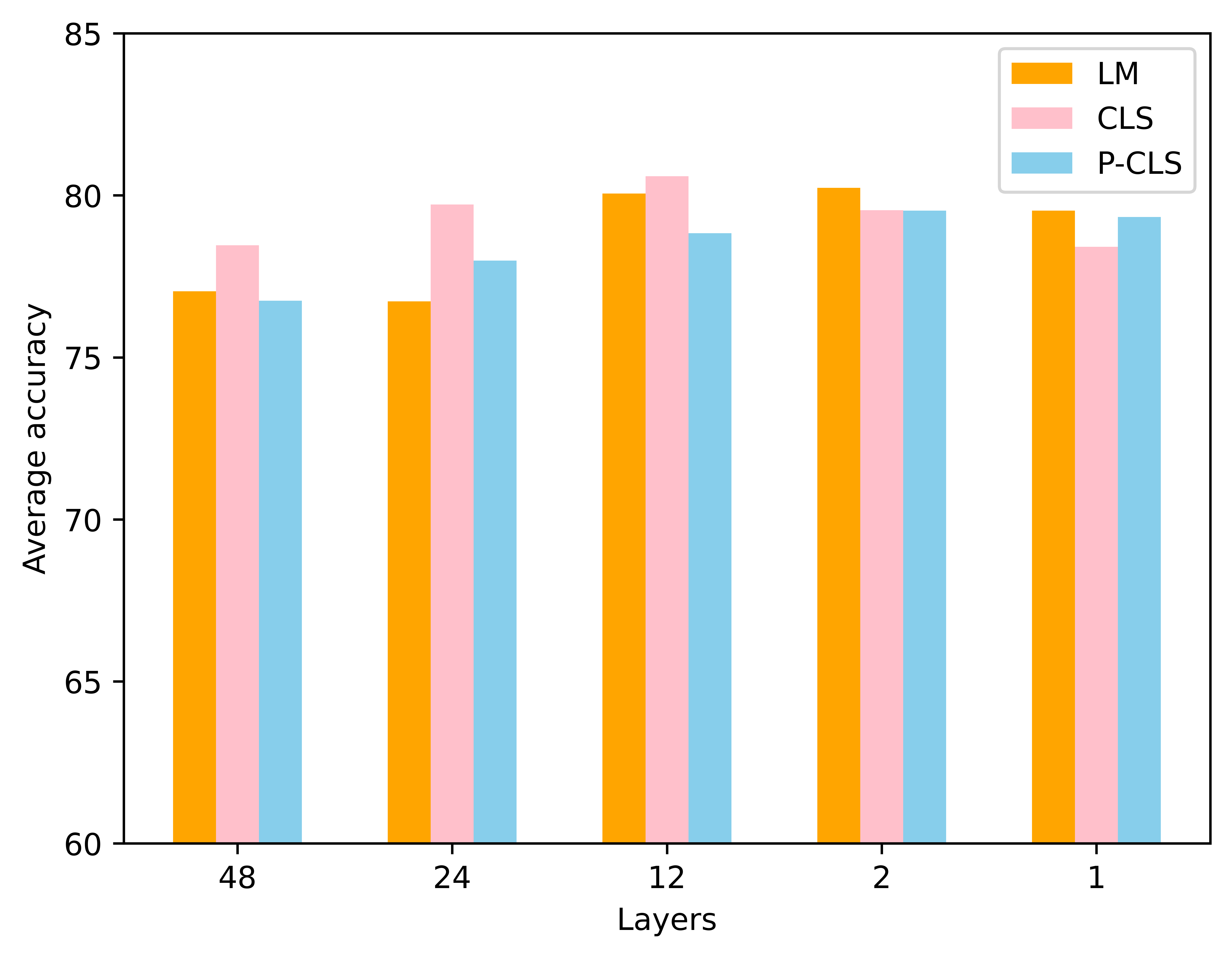} 
		\end{minipage}
	}
        \subfigure[\text{OPT}]{ 
		\begin{minipage}{\linewidth}
			\centering    
			\includegraphics[width=\linewidth]{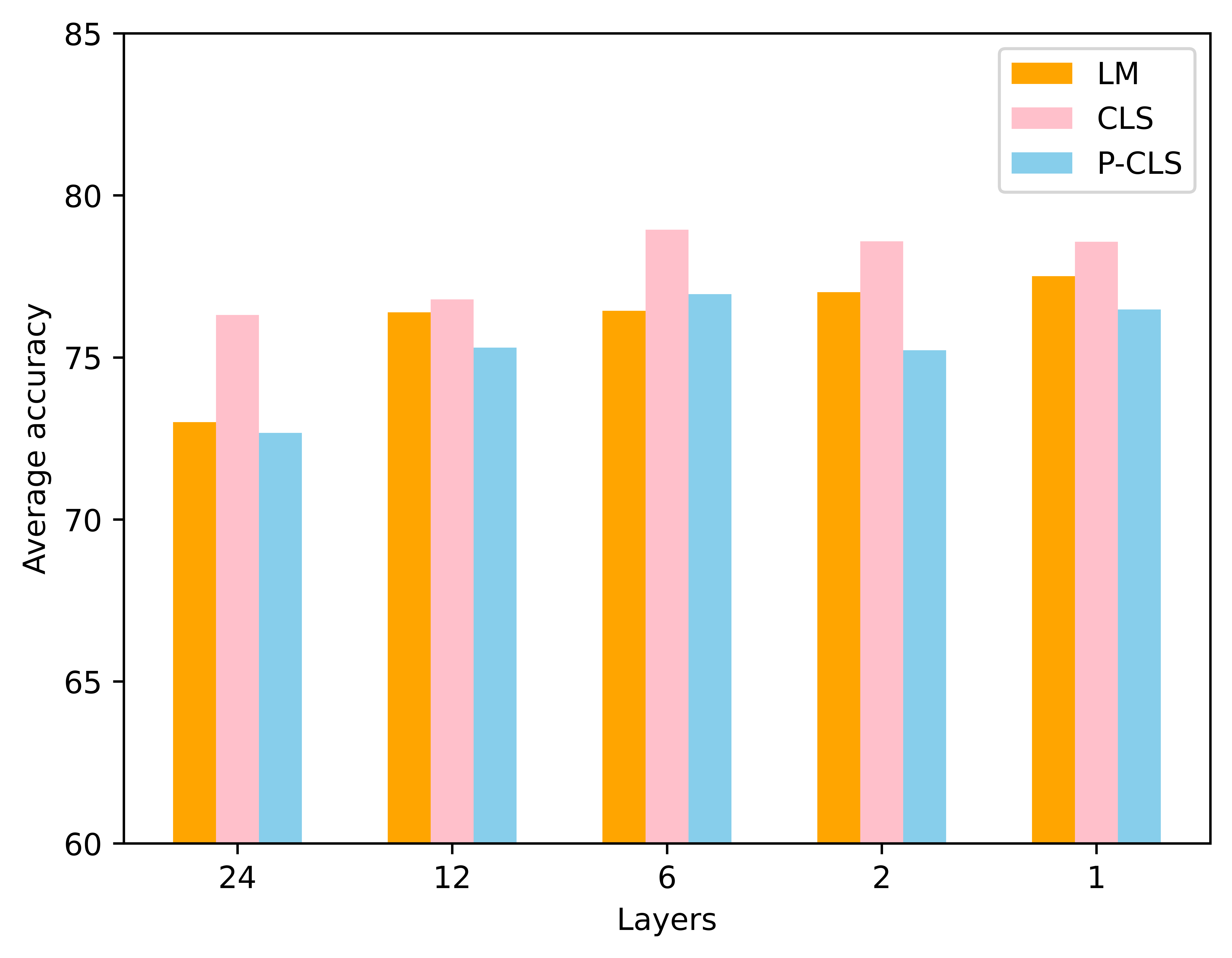}
		\end{minipage}
	}
 
		\centering\caption{The average accuracy of different layers for three classification methods. LM denotes language modeling head, P-CLS denotes prompt-based fine-tuning with classification head, CLS denotes classification head without prompt.}\label{comparison}
	\end{figure}

\subsection{Replacing Language Modeling Head}

To investigate whether the consistent performance across different layer configurations is influenced by the use of the language modeling head, we replace it with a classification head and repeat the experiments. 
The results are presented in Table \ref{cls_prompt_results}. In comparison to the language modeling head, the performance of the classification head exhibits a slight decrease under identical layer counts. Nonetheless, the average accuracy of the classification head maintains consistent performance with marginal fluctuations. 
For GPT2-XL, employing the full complement of layers yields an average accuracy of 76.75\%. Remarkably, the model achieves its highest accuracy with only 2 layers, with even a single-layer configuration outperforming the full-layer model. A similar trend is observed for the OPT model, wherein a reduction in layers correlates with performance enhancement. While the full-layer configuration achieves an average accuracy of 72.67\%, the single-layer model surpasses it with an accuracy of 76.48\%.

We visually represent the results of the three classification methods we have implemented in Figure \ref{comparison}. Notably, these methods yield similar outcomes, with layer count exerting negligible influence on model performance. Intriguingly, models with fewer layers often outperform those with full layers. These findings suggest that the consistent performance observed with fewer layers is not solely attributable to the language modeling head. Consequently, this discovery holds promise for optimizing resources in classification tasks by economizing training and storage resources.

\subsection{Discussion}

The consistent performance across different layer configurations suggests that LLMs are highly redundant. Lower layers capture general linguistic and semantic features that are sufficient for relatively simple classification tasks, while upper layers may introduce task-specific representations that are less critical in few-shot learning scenarios. The improved performance of models with fewer layers may be attributed to reduced overfitting. In few-shot learning scenarios, larger models with more parameters are more prone to overfitting, while smaller models with fewer layers may generalize better.
The parameter reductions achieved by layer pruning are substantial. For GPT-2 XL, reducing from 48 layers to 1 layer results in a 93\% reduction in parameters, while for OPT-1.3B, reducing from 24 layers to 1 layer results in an 88\% reduction. These reductions translate to significant savings in memory and computational costs, making LLMs more practical for deployment in resource-constrained environments.
The results indicate that different tasks may have varying sensitivity to layer pruning. For example, sentiment analysis tasks (e.g., SST-2) may require slightly deeper representations, while topic classification tasks (e.g., AGNews) can be effectively handled by shallow models.
The findings suggest that layer pruning can be an effective strategy for optimizing LLMs for classification tasks, particularly in scenarios where computational resources are limited. By reducing the number of layers, it is possible to achieve comparable or even better performance while significantly reducing the model size.

\section{Conclusion}
This study delves into the impact of layer count on the performance of LLMs. Our investigation unveils that variations in the number of layers do not result in performance degradation for classification tasks employing LLMs. Remarkably, even with a reduction in the number of LLM layers from 48 to 1, both GPT2-XL and OPT consistently maintain or even enhance performance. 
Our findings provide valuable insights into reducing the training and storage costs associated with LLMs, presenting promising avenues for optimizing resource allocation in such models.

\section*{Acknowledgement}

This work was supported by Center for Scalable Data Analytics and Artificial Intelligence (ScaDS.AI), German Federal Ministry of Education and Research (BMBF) via ``LLM4Edu'', a Software Campus project (01IS17042), Munich Center for Machine Learning (MCML), 
and China Scholarship Council (CSC).



\bibliographystyle{splncs04}
\bibliography{custom, anthology}

\end{document}